\documentclass{article} 
\usepackage{iclr2025_conference,times}

\usepackage{amsmath,amsfonts,bm}

\def\eqref#1{equation~\ref{#1}}

\def\1{\bm{1}}

\DeclareMathAlphabet{\mathsfit}{\encodingdefault}{\sfdefault}{m}{sl}
\SetMathAlphabet{\mathsfit}{bold}{\encodingdefault}{\sfdefault}{bx}{n}

\def\gA{{\mathcal{A}}}

\def\gC{{\mathcal{C}}}
\def\gD{{\mathcal{D}}}
\def\gE{{\mathcal{E}}}

\def\gK{{\mathcal{K}}}

\def\gT{{\mathcal{T}}}

\def\gX{{\mathcal{X}}}
\def\gY{{\mathcal{Y}}}

\usepackage{multirow}
\usepackage{hyperref}
\usepackage{booktabs} 
\usepackage{url}
\usepackage{pdflscape}
\usepackage{subfigure}
\usepackage{graphicx}
\usepackage{makecell}
\usepackage{algorithm}
\usepackage{algorithmic}
\usepackage{amsmath}
\usepackage{wrapfig}
\usepackage{svg}
\usepackage{listings}
\usepackage[most]{tcolorbox}
\usepackage{xcolor}
 \usepackage{amssymb}
\definecolor{c1}{RGB}{239,118,122}
\definecolor{c2}{RGB}{69,105,144}
\definecolor{c3}{RGB}{72,192,170}
\definecolor{c4}{RGB}{179,149,189}

\title{MetaAgent: Toward Self-Evolving Agent via Tool Meta-Learning
}

\iclrfinalcopy
\author{%
  Hongjin Qian, Zheng Liu\thanks{Corresponding Author} \\
  BAAI\\
  \texttt{\{chienqhj, zhengliu1026\}@gmail.com} 
}

\begin{document}

\maketitle

\begin{abstract}

In this work, we propose MetaAgent, an agentic paradigm inspired by the principle of learning-by-doing, where expertise is developed through hands-on practice and continual self-improvement. MetaAgent starts with a minimal workflow, equipped only with basic reasoning and adaptive help-seeking abilities. When a knowledge gap is encountered, MetaAgent generates natural language help requests, which are routed to the most suitable external tool by a dedicated tool router. As MetaAgent solves tasks, it continually conducts self-reflection and answer verification, distilling actionable experience into concise texts that are dynamically incorporated into future task contexts. Besides, MetaAgent autonomously builds in-house tools and a persistent knowledge base by organizing its tool-use history, further enhancing its ability to retrieve and integrate relevant information
We term this continual, data-driven process as \textit{meta tool learning}, through which MetaAgent incrementally refines its reasoning and tool-use strategies, without changing model parameters or requiring further post-training. Evaluated on challenging knowledge discovery benchmarks, including GAIA, WebWalkerQA, and BrowseComp, MetaAgent consistently outperforms workflow-based baselines and matches or exceeds end-to-end trained agents, demonstrating the promise of self-evolving agentic systems for robust, general-purpose knowledge discovery. We provide our source codes in \href{https://github.com/qhjqhj00/MetaAgent}{\textit{this repository}}.
\end{abstract}

\section{Introduction}
Recent information-seeking systems powered by large language models (LLMs), such as ChatGPT, have achieved remarkable success~\citep{chatgpt,gemini}. These systems can provide instant answers to almost any type of information need, ranging from simple factual queries to complex information aggregation~\citep{zhao2024retrievalaugmentedgenerationrag,agentsurvey}. However, when faced with more complex tasks that require synthesizing information across multi-step reasoning or interacting with external tools, most current LLMs struggle to deliver accurate solutions, especially in deep knowledge discovery scenarios~\citep{li2025towards, HLE,zhu2024largelanguagemodelsinformation}. 

\begin{figure}[t]
    \centering
    \includegraphics[width=0.9\linewidth]{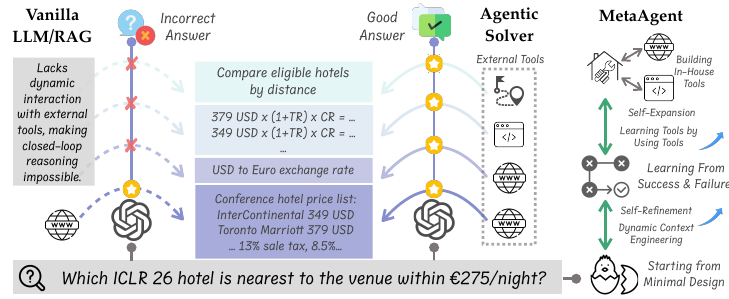}
    \caption{Illustration of a deep knowledge discovery task with sequential information dependencies, requiring multi-step reasoning and dynamic tool use to arrive at the correct answer. These tasks challenge standard LLMs and RAG approaches, which lack flexible tool interaction. Agentic systems address this limitation. MetaAgent demonstrates this approach by starting with a minimal, adaptable design and evolving through continual, data-driven task completion.}
    \label{fig:idea}
\end{figure}

As illustrated in Figure~\ref{fig:idea}, the query “Which ICLR 26 hotel is nearest to the venue within €275/night?” exemplifies a simplified deep knowledge discovery scenario. To answer this question, an agent must first identify the list of conference hotels through web search, then check current currency exchange rates. By accurately converting prices to the target currency using computational tools and comparing the distances of eligible hotels to the venue, the agent can determine the correct answer. However, even when equipped with search tools, standard LLMs struggle with such queries because they cannot effectively manage the sequential reasoning and tool use required to integrate information from multiple sources and execute each step of the problem~\citep{qian2025hawkbench,wei2025browsecomp,zhao2024retrievalaugmentedgenerationrag}.

To overcome such challenges, recent advances in agentic AI have focused on harnessing the reasoning capabilities of LLMs not only for generating answers, but also for planning and executing multi-step solutions to complex tasks, enabling scalable deep reasoning even at test time~\citep{deepresearch,snell2024scaling}. In this agentic paradigm, the LLM acts as a central coordinator: it decomposes a complex query into manageable sub-tasks, interacts with external tools, such as search engines or code compilers, and incrementally integrates the information obtained at each step into its reasoning trajectory~\citep{searcho1,searchr1,li2025webthinker,deepresearcher,dong2025tool}. By following this paradigm, agentic systems can resolve intricate dependencies and deliver accurate solutions to deep knowledge discovery problems that static LLMs alone cannot handle~\citep{li2025towards}.

To implement agentic systems, there are currently two main approaches. The first approach is to manually design task-specific workflows, where human experts predefine how the agent should plan tasks and use tools~\citep{searcho1,wu2025agentic,soni2025coding}. Although this method can be effective for narrow applications, it relies heavily on human expertise and lacks the flexibility needed to adapt to diverse deep knowledge discovery scenarios~\citep{li2025towards}. The second approach is to train LLMs end-to-end to perform multi-step reasoning and tool use, encouraging the model to explore effective reasoning trajectories with optimization methods such as reinforcement learning~\citep{searchr1,li2025webthinker,wang2025actingreasoningmoreteaching}. This enables the model to autonomously decompose complex tasks into atomic sub-tasks and interact with external tools during inference. Although this method offers stronger generalization, it requires large amounts of annotated, task-specific data, which are often difficult and costly to obtain~\citep{jin2025empirical,zhang2025processvsoutcomereward,wu2025webdancerautonomousinformationseeking}. Additionally, once the model is fine-tuned for specific tasks, its performance on other tasks may decline, and it can inherit biases from the training data~\citep{zhang2024agentic}.

In this work, we propose MetaAgent, an alternative agentic paradigm that begins with a minimal yet highly generalizable workflow and evolves naturally through data-driven task completion. Inspired by how humans progress from novices to experts by accumulating experience, MetaAgent is initialized with only two core abilities: autonomous reasoning and adaptive help-seeking. For any complex task, it reasons step by step using its own knowledge, and upon reaching the limits of its understanding, generates help requests to seek external assistance. These help requests are handled by a dedicated tool router, which flexibly maps each request to the most suitable external tool for execution. This design allows MetaAgent to address a wide range of information-seeking and deep knowledge discovery scenarios, while keeping its operational core simple, adaptable, and scalable.

Beyond its minimal initialization, MetaAgent continually evolves through a process grounded in metacognitive theory~\citep{schraw1995metacognitive}, which we term meta tool learning. After each task, MetaAgent actively engages in self-reflection and answer verification—core practices of metacognitive self-regulation in cognitive science. During this process, the agent not only reviews the accuracy of its answers, but also systematically examines its reasoning patterns, evaluates the effectiveness of tool selection and usage, and considers how external information was integrated. By identifying recurring cognitive biases and generalizable pitfalls, as well as distilling successful decision-making strategies, MetaAgent abstracts broadly applicable insights from each experience. These distilled learnings are then dynamically incorporated into subsequent input contexts, enabling MetaAgent to continually refine its planning, reasoning, and tool-use strategies as it encounters new tasks. Meanwhile, by organizing its history of interactions with the tool router, MetaAgent constructs an in-house knowledge base that grows more informative over time, supporting efficient retrieval and integration of external information. Through this combination of dynamic context engineering and continual knowledge acquisition, MetaAgent achieves robust, data-driven adaptation to complex and evolving tasks, without updating model parameters or relying on large-scale post-training.

We evaluate MetaAgent on three challenging deep knowledge discovery benchmarks, including General AI Assistant (GAIA)~\citep{mialon2023gaia}, WebWalkerQA~\citep{wu2025webwalker}, and BrowseComp~\citep{wei2025browsecomp}. Experimental results demonstrate that, with the same LLM backbone, MetaAgent consistently outperforms well-crafted workflow-based baselines across all datasets and achieves competitive or superior performance compared to end-to-end trained agentic systems. The main contributions of this paper are as follows:
(1)~We introduce the concept of meta tool learning, a continual, data-driven process that enables agentic systems to incrementally acquire task planning and tool-use strategies, without relying on manual workflow design or large-scale end-to-end post-training.
(2)~We present MetaAgent, a self-evolving agentic system that starts from a minimal, generalizable workflow and progressively enhances its reasoning and tool-use abilities via self-reflection, verified reflection, dynamic context engineering, and in-house tool construction.
(3)~Through comprehensive experiments, we show that MetaAgent consistently outperforms workflow-based baselines and matches or surpasses the performance of end-to-end trained agents, highlighting the promise of self-evolving agents for robust and generalizable knowledge discovery.

\section{Method}
\begin{figure*}[t]
    \centering
    \includegraphics[width=\linewidth]{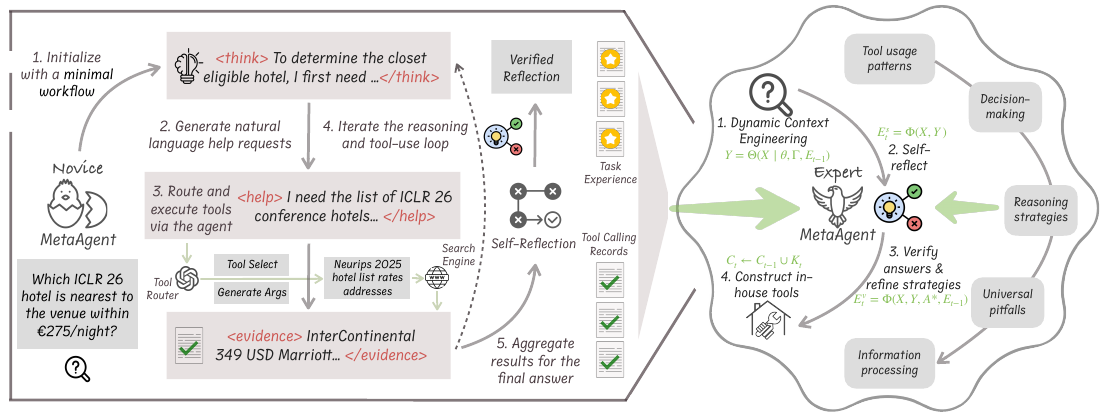}
    \caption{Overview of MetaAgent. The left side illustrates MetaAgent’s minimal workflow, where records of tool use and task experiences are accumulated to inform future reasoning. The right side depicts MetaAgent’s evolution: it continually leverages these accumulated experiences and enriches its in-house tools, enabling progressive self-improvement. Through this process of tool meta learning, MetaAgent advances from novice performance to expert-level capabilities.}
    \label{fig:model}
\end{figure*}

\subsection{Preliminary}
The process of information seeking with LLMs can be generally formulated as $\gY = \Theta(\gX)$, where $\gX$ is the input sequence, typically including a user query, task instruction, and other relevant context. $\gY$ is the model’s output. For simple queries, this direct mapping often suffices. However, in complex knowledge discovery scenarios, LLMs often cannot generate accurate answers relying solely on their internal knowledge. In such cases, they must leverage external tools $\gT$—such as web search engines, local retrievers, or code executors—to obtain supplementary information $\gK$ that compensates for their limitations and enables them to arrive at correct solutions.

A straightforward and widely used strategy for integrating external information is to directly inject all available knowledge $\gK$ into the input sequence $\gX$, effectively treating external content as a static extension of the prompt. For tasks requiring multiple rounds of tool use, this paradigm is often instantiated as a multi-agent or pipeline-based system: each specialized agent $\Theta_i$ handles a specific sub-task, with its outputs and any newly acquired external knowledge $\gK_i$ sequentially appended to the context and passed on to the next agent in the pipeline, which can be formalized as:
\begin{align}
\gY = \Theta_N\Big(\ldots, \Theta_2\big(\Theta_1(\gX_1) \cup \gK_1\big) \cup \gK_2, \ldots \cup \gK_{N-1}\Big),
\label{eq:workflow}
\end{align}
where $\gX_1$ denotes the initial input, and $\gK_i$ represents the external knowledge injected at each step. While this design is straightforward, it often lacks global awareness and flexibility, since each intermediate result must be explicitly incorporated into the next stage’s input. This not only reduces efficiency and scalability for complex knowledge discovery, but also hinders joint optimization, as reasoning and coordination are fragmented across multiple agents.

Recent advances in test-time scaling have led to a more dynamic paradigm for tool use. Instead of having multiple agents handle different sub-tasks in sequence, a single central agent now takes charge of the full reasoning and tool-calling process during one inference. Rather than preloading all external knowledge, the agent switches between thinking and calling tools step by step, adding tool results directly into its reasoning path. This unified approach is more efficient, easier to optimize, and allows for flexible, context-aware planning at each stage. Compared to static pipeline systems, it better supports complex knowledge discovery tasks by making the entire process adaptive and streamlined.
Formally, the agent’s solution trajectory can be represented as:
\begin{align}
\gY = \left( y_1, k_1, y_2, k_2, \ldots, y_T, \gA \right),
\label{eq:agent}
\end{align}
where each $y_t$ denotes an internal reasoning step and each $k_t$ is an optional external tool result (which can be empty if no tool is called at that step). At each stage, the agent decides whether to continue reasoning or invoke a tool, according to:
\begin{align}
y_t = \Theta(\gX, \gY_{<t} \mid \theta),
\end{align}
\[
k_t =
\begin{cases}
\gT(y_t) & \text{if a tool call is triggered at step $t$,} \\
\varnothing & \text{otherwise}.
\end{cases}
\]
Here, $\Theta(\cdot)$ (parameterized by $\theta$) is the central agent governing both reasoning and tool-use, while $\gT$ is the interface to external tools. This process allows the agent to adaptively and seamlessly interleave reasoning with tool use, ultimately producing the final answer $\gA$.

Compared to the pipeline-based formulation in Eq.~(\ref{eq:workflow}), which can be rigid and inefficient, the unified agentic process in Eq.~(\ref{eq:agent}) concentrates planning and execution in a single inference process. This shift places greater emphasis on developing strong internal scheduling and tool-calling strategies, which are essential for robust performance on deep knowledge discovery tasks.

\subsection{The Proposed Method: MetaAgent}

In this work, we propose MetaAgent—an alternative agentic paradigm that begins with a minimal yet highly generalizable workflow, evolving naturally through data-driven task completion. As shown in Figure~\ref{fig:model}, inspired by the way humans progress from novices to experts via accumulated experience, MetaAgent is initially equipped with two core abilities: autonomous task reasoning and adaptive help-seeking. As MetaAgent tackles diverse tasks, it continually reflects on its reasoning process and outcomes, systematically accumulating practical knowledge. In particular, MetaAgent improves its tool-use skills through repeated, hands-on interactions with external tools, progressively mastering when and how to seek and apply external assistance. This ongoing process, which we term meta tool learning, enables MetaAgent to incrementally refine both its problem-solving strategies and its ability to select and leverage the most effective tools for new and complex challenges.

\textbf{Foundation: Minimal Workflow Design. }
MetaAgent is initialized with a minimal design which follows a simple workflow:
\begin{enumerate}
\item Reason through the task using current knowledge and the available context;
\item Upon reaching its ability limitation, \emph{generate a natural language help request} $t$ describing the unmet need which would be satisfied by a stand-alone tool router agent $\Gamma$;
\item Once sufficient information is gathered, aggregate the evidence and produce the final answer $\gA$.
\end{enumerate}
MetaAgent iteratively executes this workflow within its reasoning trajectory~$\gY$, interleaving autonomous reasoning steps $y_i$, help requests $t_i$, and external knowledge $k_i$, which can be formalized as:
\begin{align}
\gY = \big( y_1, t_1, k_1, y_2, t_2, k_2, \ldots, y_n, \gA \big) = \Theta(\gX \mid \theta, \Gamma).
\label{eq:def}
\end{align}
Note that MetaAgent adopts a modular design that separates task reasoning from tool execution. The central agent’s job is to concentrate on reasoning about the problem and generating help requests in natural language, without needing to know any details about which tools are available or how to use them.

For each help request~$t_i$, MetaAgent relies on a dedicated tool router agent~$\Gamma$, which determines the most suitable tool from a set of tool descriptions~$\gD$. The tool router interprets the request and matches it to the best tool~$\gT^*_i$, which then produces tool results as the external knowledge $k_i$:
\begin{align}
k_i = \gT^*_i(\Gamma(t_i \mid \gD)).
\end{align}
Here, the tool description set~$\gD$ is designed to be easy to configure and extend, supporting flexible integration of new tools as needed.

For MetaAgent, the tool router routes to two core tools: (i)~web search tool: mapping help requests to web search queries—retrieving search results via search engines, crawling relevant web pages, and summarizing key information as external knowledge. (ii)~code executor: mapping help requests into executable code snippets, and returning executed results as external knowledge. 
At each step, the acquired external knowledge $k_i$ is incorporated back into the context, allowing the agent to dynamically refine its reasoning based on newly gathered information.

\textbf{Self-Evolving: Meta Tool Learning.}
From Eq.~(\ref{eq:def}), we observe that, with model parameters $\theta$ fixed, the quality of the reasoning trajectory ultimately depends on two factors: (i) the content and structure of the input sequence $\gX$, and (ii) the interaction protocol between the central agent and the tool router agent $\Gamma$.
To reach an informative input sequence $\gX$, previous methods typically inject expert-curated strategies or handcrafted workflows into the task instruction—a practice often referred to as prompt engineering. However, for complex knowledge discovery tasks, such predefined strategies are rarely optimal, as they fail to adapt to the nuanced requirements that emerge in real-world problem solving.

In addition, since the tool router agent $\Gamma$ functions as a standalone module that translates natural language requests into concrete tool executions, the central agent must learn to articulate its information needs precisely and effectively. This process requires ongoing refinement in both task decomposition and the communication of tool-use intent, enabling more dynamic and context-aware interactions throughout the reasoning trajectory.

To bridge the gap between novice and expert tool use, MetaAgent conducts self reflection and verified reflection after each task, distilling actionable experience into concise texts. These are dynamically incorporated into future input sequences~$\gX$, allowing MetaAgent to continually refine its task planning and tool-use strategies. Meanwhile, MetaAgent organizes its interaction history with the tool router~$\Gamma$, gradually building an in-house tool that manages accumulated external knowledge in a persistent memory. As MetaAgent interacts with external tools over time, this in-house tool becomes increasingly informative, enabling more effective retrieval and integration of relevant knowledge for future tasks.

We refer to this continual process of learning from both experience and tool use as \textbf{meta tool learning}. Through meta tool learning, MetaAgent evolves autonomously, incrementally improving its task planning and tool-use strategies without changing model parameters—empowering robust, data-driven adaptation across diverse knowledge discovery scenarios.

\paragraph{Self-Refinement: Dynamic Context Engineering}
While processing tasks, MetaAgent dynamically constructs its input sequence as $\gX_t = \{q, p, \gE_{t-1}\}$, where $q$ denotes the task description, $p$ is the basic task instruction, and $\gE_{t-1}$ represents incrementally accumulated task experience that provides insights into task planning and tool use. MetaAgent employs two reflection strategies: self-reflection and verified reflection.

First, self-reflection is conducted based on the input sequence and reasoning trajectory $(\gX, \gY)$, without access to the ground-truth answer. During this process, MetaAgent reviews its reasoning trajectory, scrutinizing the validity and factual grounding of each step. Any uncertainty or identified flaw is summarized into feedback:
\begin{equation}
\gE_\text{self} = \Omega(\gX, \gY),
\end{equation}
where $\Omega(\cdot)$ denotes a reflection process. If MetaAgent lacks confidence in its reasoning process or the concluded answer, the self-reflection experience $\gE_\text{self}$ is incorporated into the input sequence $\gX$, prompting MetaAgent to reconsider the current task and avoid previous mistakes.

Second, if ground-truth answers are available, verified reflection compares the predicted answer $\gA_\text{pred}\in\gY$ with the ground-truth $\gA^*$, further analyzing both successes and failures. Guided by metacognitive theory, this process goes far beyond simple answer checking. At each step, the agent systematically abstracts generalizable insights—such as which reasoning strategies or decision frameworks contributed to success or failure, what universal principles governed effective tool selection and information integration, and what cognitive pitfalls or errors emerged. These distilled meta-level experiences are accumulated incrementally:
\begin{equation}
\gE_t = \Omega(\gX, \gY, \gA^*, \gE_{t-1}).
\end{equation}
Here, $\Omega(\cdot)$ represents the reflective process that transforms episodic feedback into transferable procedural knowledge. Verified reflection is only performed after the answer is finalized, and the updated experience is used exclusively in subsequent tasks to prevent information leakage. Through this metacognitive abstraction, MetaAgent continually enriches its repertoire of strategies, supporting robust self-improvement across diverse domains.

Both forms of experience, $\gE=(\gE_t, \gE_{\text{self}})$, are distilled into concise, transferable texts, incrementally updated and incorporated into future input sequences, enabling MetaAgent to continually generalize and refine its task-planning and tool-use strategies. We term this reflect-refine process \textbf{dynamic context engineering}, as it continually enhances MetaAgent’s capabilities in task planning and tool utilization. In our experiments, experience is constructed online from scratch, ensuring no information leakage during evaluation. In practical applications, these experiences can also be built using annotated data and adaptively incorporated during inference to enhance performance.

\paragraph{Self-Expansion: Building In-House Tools}

In MetaAgent, the tool router agent $\Gamma$ serves as a fixed, sandboxed proxy for mapping help requests to concrete tool executions. While this modularity ensures reliable and standardized tool use, the tool router itself does not evolve or share its full perspective with MetaAgent. Typically, for each external tool call, $\Gamma$ processes all accessible information—such as the complete set of web pages retrieved by a search engine—but only returns a distilled, task-relevant subset $k_i = \Gamma(\gK_i)$ to the central agent. Here, $\gK_i \subseteq \gK$ represents the portion of all external knowledge $\gK$ encountered in the $i$-th tool call. Over a series of $T$ steps, the information available to MetaAgent is thus:
\begin{align}
    k_{1:T} = \{k_1, k_2, \ldots, k_T\} = \{\Gamma(\gK_1), \Gamma(\gK_2), \ldots, \Gamma(\gK_T)\},
\end{align}
which forms only a filtered view of the richer, underlying information subsets $\gK_{1:T}$. This creates an inherent information gap between what the agent receives and the full spectrum of external data processed by the tool router.

To bridge this gap and enable a more holistic understanding, MetaAgent incrementally constructs an in-house tool—a persistent local knowledge base $\gC$—by accumulating the raw interaction history and all intermediate results from each tool execution. Formally, this knowledge base is updated as:
\begin{align}
    \gC \leftarrow \gC \cup \left(\gK_1 \cup \gK_2 \cup \cdots \cup \gK_T \right).
\end{align}
Unlike the fixed outputs $k_i$ returned by $\Gamma$, the in-house tool $\gC$ provides MetaAgent with global access to the accumulated external knowledge across the entire reasoning episode. As the agent interacts with external tools over time and across tasks, $\gC$ becomes progressively richer and more informative, empowering MetaAgent to (i) revisit previous tool executions, (ii) cross-reference retrieved evidence, and (iii) develop memory-based retrieval or summarization utilities tailored to its own evolving needs.

Given a batch of input tasks, MetaAgent can first initialize the local knowledge base $\gC$ by simulating its reasoning trajectories, proactively gathering and storing sufficient external knowledge through the tool router agent. Once this knowledge base is constructed, MetaAgent proceeds to solve these tasks, now grounded by both the in-house tool $\gC$ and the standard outputs from the tool router agent $\Gamma$. This two-stage process further enhances MetaAgent’s ability to integrate global context, enabling more robust and informed reasoning across diverse deep knowledge discovery tasks.
By integrating all of MetaAgent’s design components, Eq.(\ref{eq:def}) can be reformulated as:
\begin{align}
\gA \in \gY = \Theta(\gX \mid \theta, \Gamma, \gE, \gC),
\end{align}
indicating that MetaAgent produces the answer$\gA$ for a given task based on the central agent’s reasoning, supported by the tool router~$\Gamma$, accumulated experience~$\gE$, and the in-house knowledge base~$\gC$.

\section{Experiment}
\subsection{Evaluation Settings}
\paragraph{Baselines}
We benchmark MetaAgent against four categories of baselines:
(1) Direct Reasoning: strong LLMs such as Qwen2.5-32B, QwQ-32B, GPT-4o, and DeepSeek-R1-671B~\citep{deepseekr1,qwen2025qwen25technicalreport,gpt-4}, evaluated without external tool access;
(2) Direct Reasoning with Retrieval: LLMs enhanced with standard retrieval augmentation, injecting retrieved context into the input;
(3) Expert-Designed Workflows: autonomous agents guided by manually crafted planning and tool-use strategies, including Query Planning, Interactive Task Planning, ReAct-style agents, and the expert-curated Search-o1~\citep{searcho1,yao2022react,ITER-RETGEN};
(4) End-to-End Trained Agents: systems such as WebThinker~\citep{li2025webthinker}, which are optimized for autonomous search and tool interaction through end-to-end training.
To ensure fairness, QwQ-32B and Qwen2.5-32B are used as the backbone LLMs for most baselines.

\paragraph{Datasets}
We evaluate MetaAgent and baseline methods on three challenging deep knowledge discovery benchmarks, including:

\textbf{(1)~GAIA (General AI Assistant)} is a benchmark of over 450 real-world questions that test a range of capabilities including multi-step reasoning, multi-modal understanding, web browsing, and tool use~\citep{mialon2023gaia}. It is divided into three levels of increasing difficulty, with human accuracy at ~92\% while advanced LLM agents (e.g., GPT‑4 with plugins) score around ~15\%~\citep{mialon2023gaia,zhao2024surveylargelanguagemodels}. GAIA is particularly valuable for evaluating general-purpose agentic reasoning and tool-using proficiency. Following previous work~\citep{li2025webthinker, wu2025webdancerautonomousinformationseeking}, we use 103 questions from the text-only validation subset.
\textbf{(2)~WebWalkerQA} evaluates an agent’s ability to perform structured web traversal to find complex, multi-source answers~\citep{wu2025webwalker}. It contains 680 queries derived from real-world scenarios across domains like conferences and organizations. Agents must navigate through subpages and synthesize information, making WebWalkerQA a challenging testbed for long-horizon tool-enabled reasoning~\citep{wu2025webwalker}.
\textbf{(3)~BrowseComp} is a browsing-centric benchmark containing 1,266 “hard-to-find” questions that test agents’ persistence and creativity in web navigation~\citep{wei2025browsecomp}. While answers are short and easily verifiable, they are intentionally crafted to avoid the top search results, forcing agents to explore deeply across numerous web pages. This benchmark emphasizes the need for effective multi-step planning under uncertainty. Given the extreme difficulty—often requiring agents to browse hundreds of pages per question—we evaluate on two selected topics, \textit{Art} and \textit{History}, totaling 257 questions.

Each of these datasets stresses different aspects of deep knowledge discovery—from tool use and multi-step planning (GAIA), to website traversal (WebWalkerQA), to long-horizon, creative browsing (BrowseComp)—making them complementary evaluations of MetaAgent’s self-evolving reasoning and tool-integration abilities.

\paragraph{Implementation Details}
In our main experiments, MetaAgent employs QwQ-32B as the central reasoning agent. For web search, we use Google’s \href{https://developers.google.com/custom-search/v1/overview}{Custom Search JSON API}, and Jina AI’s \href{https://jina.ai/api-dashboard/reader}{web page reader} is utilized to retrieve web content. Additionally, we employ Python’s \texttt{exec()} function to execute code snippets and perform calculations.
To prevent information leakage, task reflection in MetaAgent is conducted unilaterally: the summarized experience from each task is used only for subsequent tasks, never for the current one.
For the in-house tool, MetaAgent uses BGE-m3 embeddings to index all browsed web content as a persistent knowledge memory~\citep{bge_m3}. To ensure a solid knowledge foundation, we initialize the in-house tool by simulating task completion three times for each benchmark, thereby proactively gathering and constructing the persistent memory required by MetaAgent.
For all baselines, we either report the results from original papers or run the official implementations. All experiments are conducted on eight NVIDIA A100-40G GPUs, and MetaAgent is implemented using the \href{https://www.langchain.com/langgraph}{LangGraph} framework.

\begin{table*}[t]
    \centering
    \scriptsize
    \caption{Main experimental results. The best results are highlighted in \textbf{bold}, and the second-best are \underline{underlined}. Following these benchmark's official settings, we report Exact Match (EM) scores for the General AI Assistant and BrowseComp benchmarks, and the LLM equivalence accuracy for WebWalkerQA.}    
\begin{tabular}{lp{.75cm}p{.75cm}p{.75cm}p{.4cm}p{.4cm}p{.5cm}p{.4cm}p{.4cm}p{.4cm}p{.5cm}p{.4cm}}
\toprule
\multirow{2}[2]{*}{\textbf{Method}}& \multicolumn{4}{c}{\textbf{General AI Assistant}} & \multicolumn{4}{c}{\textbf{WebWalkerQA}}&\multicolumn{3}{c}{\textbf{BrowseComp}}  \\
  & Level 1 & Level 2 & Level 3 & Avg. & Easy & Medium & Hard & Avg.& Art & History  & Avg.\\

\midrule
\multicolumn{12}{l}{\textit{\textbf{Direct Reasoning (w/o Retrieval)}}} \\
Qwen2.5-32B & 20.5 & 9.6 & 8.3 & 13.6 & 3.8 & 2.5 & 3.3 & 3.1    & 0.0 & 0.0 & 0.0   \\
QwQ-32B   & 30.8 & 15.4 & \textbf{25.0} & 22.3 & 7.5 & 2.1 & 4.6 & 4.3    & 0.0 & 0.0 & 0.0   \\
GPT-4o  &   23.1 & 15.4 & 8.3 & 17.5 & 6.7 & 6.0 & 4.2 & 5.5  & 0.8 & 0.8 & 0.8  \\
DeepSeek-R1-671B  & 43.6 & 26.9 & 8.3 & 31.1 & 5.0 & 11.8 & 11.3 & 10.0 & 0.0 & 0.0 & 0.0   \\
\hline
\multicolumn{12}{l}{\textit{\textbf{Direct Reasoning (w/ Retrieval)}}} \\
RAG (Qwen2.5-32B) & 12.8 & 11.8 & 8.3 & 11.8 & 23.1 & 14.3 & 11.3 & 15.3  & 0.0 & 0.0 & 0.0  \\
RAG (QwQ-32B)    & 33.3 & 36.5 & 8.3 & 32.0 & 36.9 & 26.1 & 33.5 & 31.2 & 0.0 & 0.0 & 0.0    \\

\midrule
\multicolumn{12}{l}{\textit{\textbf{Workflow-Based Agent}}} \\

Query Planning (Qwen2.5-32B)  & 30.8 & 17.3 & 0.0 & 20.4 & 29.4 & 36.4 & 25.0 & 30.7& 0.0 & 0.0 & 0.0 \\
Query Planning  (QwQ-32B) &  48.7 & 25.0  & 8.3 & 32.0 & 28.8 & 35.7 & 30.8 & 32.5& 0.0 & 0.8 & 0.4\\

Iterative RAG  (Qwen2.5-32B)   &  35.9 &  19.2 &  8.3 & 24.3 & 30.6 & 35.7 & 25.4 & 30.9 & 0.0 & 0.0 & 0.0\\
Iterative RAG (QwQ-32B)  & 51.3 & 28.8  & 8.3 & 35.0 & 29.4 & 32.9 & 31.3 &31.5 & 0.8 & 0.0 & 0.4 \\

ReAct (Qwen2.5-32B)  & 46.1 &\underline{44.2} & 8.3 & 40.7 & 44.3 & \underline{46.7} & 29.2 & 38.4 & 0.0 & 0.0 & 0.0\\
ReAct (QwQ-32B) & 48.7 & 34.6 & \underline{16.7} & 37.8 & 35.6 & 29.1 & 13.2 & 24.1 & 0.8 & 0.8 & 0.8\\
ReAct (GPT-4o) & 51.2  & 34.6 & 8.3 & 34.6 & 34.6 & 42.0 &  23.9 & 33.8 & \underline{2.4} & 1.6 & 1.9\\

Search-o1-32B  & 53.8 &\underline{44.2} & \underline{16.7} & 39.8 & 43.1 & 35.0 & 27.1 & 34.1  &1.6 & 2.4 & 1.9   \\
\midrule
\multicolumn{12}{l}{\textit{\textbf{End-to-end optimized Agent}}} \\

WebThinker-32B-Base & 53.8 &\underline{44.2} & \underline{16.7} & 44.7 & 47.5 & 41.1 & 39.2 & 41.9 & \underline{2.4} & 2.4 & 2.3  \\
WebThinker-32B-RL  & 56.4 & \textbf{50.0} & \underline{16.7} & \textbf{48.5} & \textbf{58.8} & 44.6 & \underline{40.4} & \underline{46.5} & \underline{2.4} & \underline{3.1} & \underline{2.7}\\
\midrule
\multicolumn{12}{l}{\textit{\textbf{Our method}}} \\
\textbf{MetaAgent}~(QwQ-32B) & \textbf{61.5} & 42.3 & \textbf{25.0} & \underline{47.6} & \underline{55.0} & \textbf{49.6} & \textbf{47.9} & \textbf{52.1} & \textbf{7.9} & \textbf{6.4} & \textbf{7.1} \\
\bottomrule
\end{tabular}
\label{tab:exp}
\end{table*}

\subsection{Main Results}

Table~\ref{tab:exp} presents the main experimental results. Overall, MetaAgent consistently outperforms expert-guided workflow baselines and achieves competitive or even superior results compared to end-to-end trained methods. Specifically: (1)~On the General AI Assistant benchmark, which requires invoking various external tools, prior methods typically enumerate all tools in the task instruction. In contrast, MetaAgent simply issues help requests and relies on the tool router for tool selection. This modular design allows MetaAgent to focus on core reasoning, resulting in improved performance. (2)~For WebWalkerQA, which centers on web browsing tasks, the agent primarily needs to search and navigate web information. Because these actions are highly homogeneous across tasks, MetaAgent’s meta tool learning enables effective transfer of experience, allowing it to start with minimal guidance yet attain strong performance. (3)~BrowseComp poses an especially difficult challenge, of which many tasks require exhaustively browsing up to hundreds of candidate web pages. In this setting, a single search-refine loop rarely uncovers the full knowledge space needed. Here, MetaAgent’s persistent in-house tool—an evolving memory of previously browsed content—offers a distinct advantage, empowering the agent to address the most demanding information aggregation challenges.
\subsection{Discussions}

\paragraph{Abalation Study}

\begin{figure}[t]
    \centering
    \includegraphics[width=0.6\linewidth]{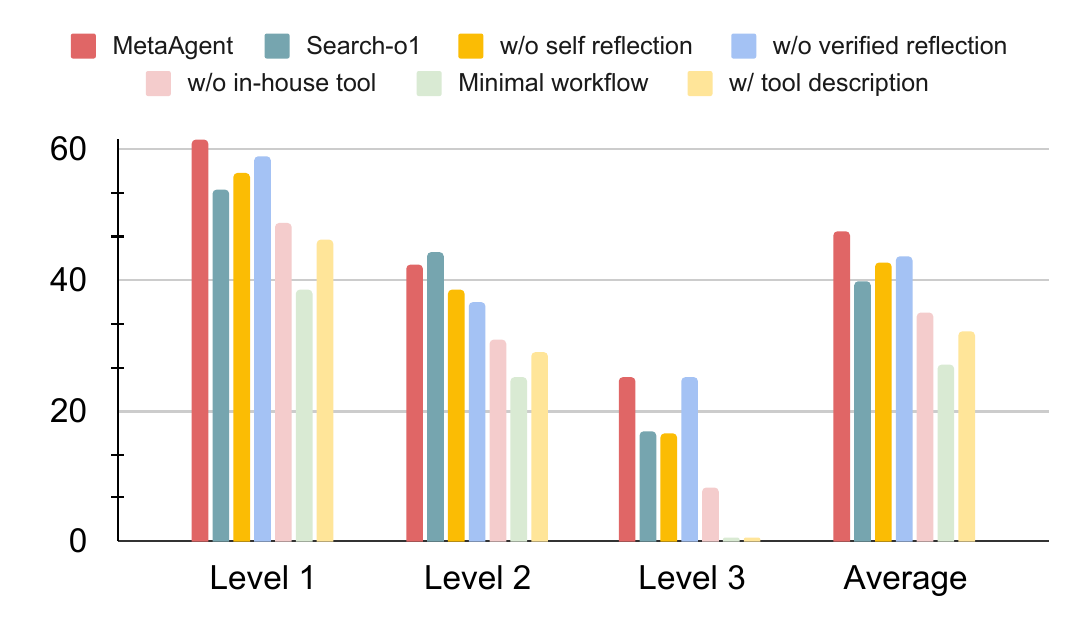}
    \caption{Ablation Study. Key components of MetaAgent are individually removed and evaluated on the GAIA benchmark.}
    \label{fig:abl}
\end{figure}

We conduct an ablation study to assess the effectiveness of MetaAgent’s modular design. Specifically, we evaluate five ablated variants: (1) removing self reflection (w/o self reflection); (2) removing verified reflection (w/o verified reflection); (3) removing the use of in-house tools (w/o in-house tool); (4) using only the minimal workflow and (5) replacing the tool router agent with direct in-context tool descriptions (w/ tool description). The results are shown in Figure~\ref{fig:abl}, which demonstrate that the full MetaAgent design enables clear progression from novice to expert behavior. Notably, the minimal workflow alone performs poorly and can hardly match expert-designed workflow Search-o1, as it only provides basic planning and help-seeking. Our main findings are as follows: (1) both self-reflection and verified reflection are critical, and their combination forms a meta tool learning loop that substantially boosts performance; (2) the in-house tool is essential for providing a global, persistent view of task knowledge, greatly mitigating information loss from single-pass search-and-refine approaches; and (3) substituting the tool router agent with in-context tool descriptions can yield better results than the minimal workflow, as it makes tool usage more explicit for the central agent—but this approach lacks the autonomous, continually evolving capabilities that distinguish MetaAgent’s design.

\paragraph{Incorporating with Commercial API}

\begin{table}[h]
    \centering
    \small
    \begin{tabular}{lcccc}
    \toprule
         & Level 1 & Level 2 & Level 3 & Avg.  \\
         \midrule
        QwQ-32B & 30.8 & 15.4 & \textbf{25.0} & 22.3\\ 
        Gemini-2.5-Flash & 12.8 & 5.7 & 0.0  & 7.8\\
        \midrule
        MetaAgent  \\
        \quad w/ QwQ-32B & 61.5 & 42.3 &\textbf{25.0} &47.6 \\
        \quad w/ Gemini-2.5-Flash & \textbf{64.1} & \textbf{46.2} & 16.7 & \textbf{49.5}\\
        \bottomrule
    \end{tabular}
    \caption{To assess MetaAgent’s adaptability to different LLMs, we substitute its core LLM with Google’s Gemini-2.5-Flash API and evaluate performance on the GAIA benchmark.}
    \label{tab:api}
\end{table}

\begin{table*}[t]

\small
    \centering
    \caption{Case study on a BrowseComp sample. MetaAgent begins by conducting broad searches, gathering candidate answers, and initially selecting one. After self-reflection reveals unmet criteria, MetaAgent attempts the task again, checking all constraints for a new candidate and ultimately producing the correct answer. For clarity, most intermediate reasoning steps are omitted.}
    \begin{tabular}{p{.98\linewidth}}
    \toprule
     \textbf{Task}: There's  \textbf{\textcolor{c3}{a building that was opened in the 2010s}} and then was  \textbf{\textcolor{c3}{closed in the 2020s, before December 2023}}. The building’s structure measures \textbf{\textcolor{c3}{15 meters in width at the base and the total length ranges between 1--3 km}} (inclusive). This building was \textbf{\textcolor{c3}{designed by a person who founded his studio in the 1990s}}. The size of the site where the building is \textbf{\textcolor{c3}{located is within 5--10 acres}} (inclusive). The parts were made in one of the countries of \textbf{\textcolor{c3}{Europe}}. As of \textbf{\textcolor{c3}{December 2023}}, can you tell me the \textbf{\textcolor{c4}{colour of the building}}?
     \quad  \textbf{Ground-truth answer}:  \textbf{\textcolor{c3}{Copper}}\\
     \textbf{Previous Task Experience}: \\
\textbf{\textcolor{c1}{Total Constraint Compliance}:}
Every clue or requirement should be treated as a mandatory filter—never omit any. Use checklists or structured approaches to ensureall constraints are satisfied before submitting an answer. \\[0.5em]

\textbf{\textcolor{c1}{Constraint-Driven Filtering}:}
Structure information retrieval and reasoning processes to reinforce all constraints simultaneously, rather than addressing them one at a time. Employ advanced filtering logic (e.g., Boolean operators) to exclude irrelevant candidates early. \\[0.5em]

\textbf{\textcolor{c1}{Cultural Depth in Symbolism}:}
When interpreting analogies or metaphors, always consider regional or contextual meanings—not just surface-level associations. Avoid projecting one’s own cultural assumptions onto ambiguous cues. \\
\midrule
  \begin{tcolorbox}[colback=gray!10, colframe=gray!50, sharp corners=south, boxrule=0.5pt, left=0mm, right=0mm, top=0mm, bottom=0mm]
\textbf{\textit{First try:}}  \textbf{\texttt{\textcolor{c4}{<think>}}} ...
\textbf{\texttt{\textcolor{c4}{<help>}}}
I need to check details about a building that matches these criteria:
1. Opened in the 2010s and closed before December 2023;
2. Measures around 15 meters wide at the base and is 1--3 kilometers long;
3. Designed by an architect or studio founded in the 1990s;
4. Located on a site of 5--10 acres;
5. Parts manufactured in Europe.
Please also confirm the official color or exterior material. \textbf{\texttt{\textcolor{c4}{</help>}}}
\end{tcolorbox} 
   \textbf{Call Help via Tool Router}: 
   \textbf{\textcolor{teal}{[search]}}: building opened 2010s closed before December 2023 
   \textbf{\textcolor{teal}{[search]}}: building 15 meters wide 1 km long \\
   \textbf{\textcolor{teal}{[search]}}: architect firm founded 1990s large building
   \textbf{\textcolor{teal}{[search]}}: building site 5 to 10 acres
   \textbf{\textcolor{teal}{[search]}}: building parts manufactured in Europe\\
\textbf{\textit{After browsing 48 web pages...}} \\
\textbf{Predicted Answer}: \textbf{\textcolor{red}{silver-gray with dark blue accents}} \\

\textbf{Self Reflection}:
The user asked for the \textcolor{red}{color} of a building matching several strict criteria.

\begin{itemize}
    \item \textbf{\textcolor{c1}{Search Result:}} The assistant identified a bridge in Shanghai as a candidate. Most criteria matched (\textcolor{teal}{opened in the 2010s}, \textcolor{teal}{closed in 2023}, \textcolor{teal}{correct dimensions}, \textcolor{teal}{architect's studio founded in the 1990s}, \textcolor{teal}{parts from Europe}).
    \item \textbf{\textcolor{c1}{Site Size Mismatch:}} \textcolor{red}{Site size} was \textcolor{orange}{8 hectares (19.76 acres)}, exceeding the \textcolor{red}{5--10 acre} requirement---this is a \textcolor{red}{critical constraint}.
\end{itemize}
\textbf{\textcolor{red}{Conclusion:}} The answer fits most criteria, but \textbf{ignores the explicit site size constraint}, which may affect correctness.\\

\begin{tcolorbox}[colback=gray!10, colframe=gray!50, sharp corners=south, boxrule=0.5pt, left=0mm, right=0mm, top=0mm, bottom=0mm]
\textbf{\textit{Second try:}}  \textbf{\texttt{\textcolor{c4}{<think>}}} ... In the last attempt, \textbf{Vessel (Hudson Yards)} was also considered as a candidate answer...
\textbf{\texttt{\textcolor{c4}{<help>}}}
I need to verify the exact dimensions, site size, closure date, and color of the Vessel structure in Hudson Yards, New York, designed by Heatherwick Studio. Specifically: 
1. Was it opened in the 2010s and closed before December 2023?
2. Does its base width measure ~15m and total length fall between 1-3km?
3. Was the site size within 5-10 acres?
4. What is its official color/material description?\textbf{\texttt{\textcolor{c4}{</help>}}}
\end{tcolorbox} 

   \textbf{Call Help via Tool Router}: 
   \textbf{\textcolor{teal}{[search]}}: Vessel Hudson Yards opening date, 
   \textbf{\textcolor{teal}{[search]}}: Vessel Hudson Yards closure date,
   \textbf{\textcolor{teal}{[search]}}: Vessel Hudson Yards base width,
   \textbf{\textcolor{teal}{[search]}}: Vessel Hudson Yards total length,
   \textbf{\textcolor{teal}{[search]}}: Vessel Hudson Yards site size acres,
   \textbf{\textcolor{teal}{[search]}}:Vessel Hudson Yards color material description \\
\textbf{\textit{After browsing 53 web pages...}} \\
\textbf{Predicted Answer}: \textbf{\textcolor{c3}{Copper}} \\
                                
\textbf{Self-Reflection} ... \textbf{\textcolor{c1}{All constraints are met}}, and the color conclusion is backed by reliable evidence. No major oversights this time.  \\

    \bottomrule
    \end{tabular}
    \label{tab:case}
   
\end{table*}

To further demonstrate MetaAgent’s flexibility and backbone-agnostic design, we substitute its central agent from QwQ-32B to Google’s Gemini-2.5-Flash API and evaluate its performance on the GAIA benchmark. As shown in Table~\ref{tab:api}, MetaAgent substantially boosts the deep knowledge discovery capabilities of both open-source and commercial LLMs, consistently outperforming their vanilla counterparts by a wide margin; this improvement is particularly notable when paired with Gemini-2.5-Flash, where MetaAgent’s self-evolving meta tool learning and dynamic context engineering unlock stronger generalization and adaptive tool-use strategies, underscoring its practical value for real-world deployments that demand seamless integration with rapidly evolving LLM backbones and commercial APIs.
\paragraph{Case Study}
\label{sec:case}
In Table~\ref{tab:case}, we present a detailed case study from BrowseComp that illustrates the full workflow of MetaAgent. The input task asks for the color of a building that satisfies multiple explicit conditions. While each individual condition may have many candidates, only their intersection leads to the unique correct answer. However, exhaustively enumerating all possibilities for every condition is non-trivial for MetaAgent. Thus, in its initial attempt, MetaAgent retrieves candidates by partially satisfying some conditions and selects a likely answer. Leveraging its self-reflection mechanism, MetaAgent then inspects whether all criteria are actually fulfilled and realizes the chosen answer is incorrect.
Informed by this self-reflection, MetaAgent launches a second attempt, this time systematically checking all constraints for another candidate. Through focused web searches, it identifies that the new candidate fully satisfies the requirements and outputs the correct answer.

Notably, the task input incorporates previously accumulated experience, strategically prompting MetaAgent to “ensure all constraints are satisfied”—a key factor in successful task completion. After solving the case, MetaAgent further reflects on its reasoning process to improve its accumulated task experience, enabling more effective handling of future tasks.

\section{Related Works}
Recent advances in LLMs have significantly improved the capabilities of AI-powered information systems~\citep{gpt-4,qwen2025qwen25technicalreport,gemini25}. Despite this progress, LLMs inherently face limitations such as knowledge staleness and hallucinations due to their static internal memories and training-data cutoffs~\citep{zhao2024surveylargelanguagemodels}. To mitigate these shortcomings, Retrieval-Augmented Generation (RAG)~\citep{lewis2020retrieval,gao2024retrievalaugmented} integrates external documents into LLM contexts. However, traditional RAG methods retrieve knowledge in a single pass with fixed queries, lacking the flexibility needed for multi-step reasoning and dynamic information integration in deep knowledge discovery tasks~\citep{zhao2024retrievalaugmentedgenerationrag,li2025towards}.

To address the limitations of static RAG, recent research has shifted toward agentic search and tool-augmented reasoning, enabling LLMs to dynamically interact with external tools during inference~\citep{WebGPT,zhang2024agentic,jin2025empirical}. These approaches empower models to iteratively reason, actively retrieve information, and refine their answers~\citep{deepresearch,deepresearcher,searchr1,li2025webthinker,wang2025chainofretrievalaugmentedgeneration}. 
Broadly, agentic systems can be categorized into two main types: workflow-based and training-based methods.

Workflow-based methods rely on carefully engineered, expert-guided strategies. For example, Mind-Map agents structure reasoning into knowledge graphs to preserve coherence over extensive reasoning chains~\citep{wu2025agentic}. Search-o1 introduces dynamic information retrieval combined with a dedicated Reason-in-Documents module to filter noisy external knowledge before integration~\citep{searcho1}. Other systems such as the generalist coding agent equip models with multiple general tools~\citep{soni2025coding}  , allowing them to perform sophisticated reasoning through modular workflows. While effective for specific applications, these methods are inherently rigid, lacking adaptability to new tasks and scenarios without substantial manual adjustments~\citep{li2025towards}.
In contrast, training-based methods leverage end-to-end training to autonomously optimize tool usage and search strategies. For instance, DeepResearcher trains agents to navigate the dynamic, unstructured web environment, employing multi-agent architectures for extracting and integrating relevant web information~\citep{deepresearcher}. Search-R1 applies reinforcement learning to teach LLMs how to generate multi-turn search queries during step-by-step reasoning, enabling interactive real-time retrieval~\citep{searchr1}. WebThinker integrates dynamic web exploration into the reasoning loop, allowing models to autonomously interleave reasoning, searching, and report drafting in an end-to-end trained fashion~\citep{li2025webthinker}. Although these approaches exhibit superior generalization compared to workflow-based methods, they rely heavily on extensive, high-quality training data and face the risk of overfitting to specific data distributions~\citep{shi2025searchrefinethinkautonomous,jin2025empirical}.

Compared to previous methods, we propose MetaAgent—an agentic system that starts from a minimal, generalizable workflow with only autonomous reasoning and adaptive help-seeking. MetaAgent evolves by reflecting on its own task-solving experiences, dynamically incorporating actionable lessons into future contexts, and gradually building an in-house knowledge base from tool interactions. This approach enables MetaAgent to progress from novice to expert through continual task completion, without relying on expert guidance or large-scale training.

\section{Conclusion}

In this paper, we introduced MetaAgent, a self-evolving agentic system designed for deep knowledge discovery with large language models. Starting from a minimal and generalizable workflow, MetaAgent is equipped only with essential reasoning and help-seeking abilities, delegating external tool interactions to a dedicated tool router agent. This simple yet flexible framework enables MetaAgent to tackle complex, multi-step tasks that require dynamic planning and the integration of external knowledge.
MetaAgent’s distinctive advantage lies in its meta tool learning paradigm. By systematically reflecting on its reasoning and tool-use strategies after each task, MetaAgent distills actionable experience into concise, transferable knowledge that is dynamically incorporated into future contexts. This enables continual adaptation and improvement—without modifying model parameters or relying on extensive post-training. Furthermore, MetaAgent incrementally builds an in-house knowledge base by organizing its interaction history, further enhancing its retrieval and integration of external information.

Comprehensive evaluations across three challenging benchmarks (GAIA, WebWalkerQA and BrowseComp) demonstrate that MetaAgent outperforms workflow-based baselines and matches or exceeds the performance of end-to-end trained agentic systems, while remaining highly generalizable and robust. These results highlight the promise of self-evolving agentic paradigms for building adaptable, scalable, and practical AI assistants capable of tackling real-world information-seeking and knowledge discovery challenges.
\bibliography{iclr2025_conference}
\bibliographystyle{iclr2025_conference}

\end{document}